\documentclass[conference]{IEEEtran}
\usepackage{times}

\usepackage[numbers,sort&compress]{natbib}
\usepackage{multicol}
\usepackage[bookmarks=true]{hyperref}
\usepackage{graphicx}
\usepackage{amsmath}

\begin{document}

\title{Deformable Cargo Transport in Microgravity \newline with Astrobee}

\author{\authorblockN{Daniel Morton}
\authorblockA{Stanford University\\
dmorton@stanford.edu}
\and
\authorblockN{Rika Antonova}
\authorblockA{University of Cambridge\\
rika.antonova@cst.cam.ac.uk}
\and
\authorblockN{Brian Coltin}
\authorblockA{NASA Ames\\
brian.coltin@nasa.gov}
\and
\authorblockN{Marco Pavone}
\authorblockA{Stanford University\\
pavone@stanford.edu}
\and
\authorblockN{Jeannette Bohg}
\authorblockA{Stanford University\\
bohg@stanford.edu}}

\maketitle

\begin{abstract}
We present \textit{pyastrobee}: a simulation environment and control stack for Astrobee in Python, with an emphasis on cargo manipulation and transport tasks. We also demonstrate preliminary success from a sampling-based MPC controller, using reduced-order models of NASA's cargo transfer bag (CTB) to control a high-order deformable finite element model. Our code is open-source, fully documented, and available at \newline \url{https://danielpmorton.github.io/pyastrobee}
\end{abstract}

\IEEEpeerreviewmaketitle

\section{Introduction}

Looking towards the future of space station logistics and maintenance, any extended uncrewed periods will have to rely on autonomous operations for tasks such as resupply and preparation of the station before/after crew arrival \cite{isaac}. In particular, to deliver supplies to the Gateway station autonomously, a robot such as Astrobee \cite{astrobee} or Robonaut \cite{robonaut} will need to transport cargo from a docked vehicle to a desired location in the station. However, the deformability of the vinyl cargo transfer bags (CTBs) makes this a difficult problem to solve. Manipulating deformable objects is an ongoing research problem \cite{deformable_manip}, primarily due to their infinite dimensionality and challenges in modeling. These challenges are further compounded by the microgravity environment: controlling an underactuated soft-robotic system often assumes a stable fixed point (for instance, hanging at rest under gravity) \cite{john_rompc, cenedese2022data}, but in microgravity, this does not exist, and the system must be actively controlled to maintain stability and prevent collision with the space station interior.

To address this, we present our preliminary work towards manipulating the deformable CTBs as Astrobee navigates through the space station. Our Python package, \textit{pyastrobee}, provides simulation, planning, modeling, and control infrastructure towards this task, and we present a sampling-based MPC which can successfully transfer the cargo between ISS modules, while avoiding collision.

\section{Simulation Environment}

\begin{figure}
  \begin{center}
    \includegraphics[width=\linewidth]{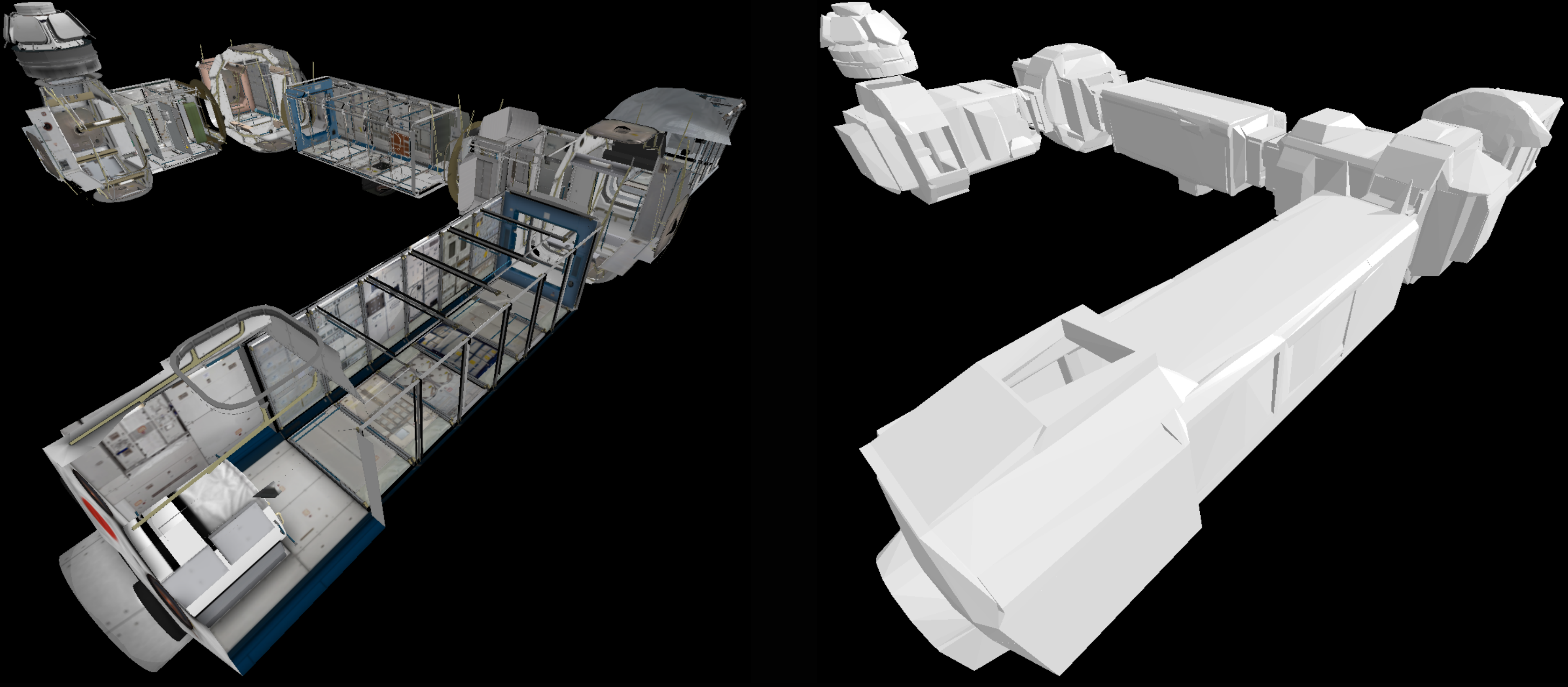}
  \end{center}
  \caption{\textbf{The ISS environment}. We build on NASA's high-quality meshes and textures for the ISS (left), and additionally provide an approximate convex-decomposition collision representation for each module (right)}
  \label{fig:env}
\end{figure}

To simulate the cargo manipulation task, we provide a realistic ISS environment, along with models of Astrobee and the CTBs. We build on Bullet physics \cite{pybullet}, which supports finite element model (FEM) deformable physics for imported surface and volumetric meshes (this feature is not supported in NASA's existing Astrobee simulation \cite{astrobee_repo}, due to limitations of Gazebo \cite{gazebo}). Our environment (Fig. \ref{fig:env}) contains high-quality visual meshes and textures from NASA's Gazebo simulation, with collision geometry created through approximate convex decomposition \cite{mamou2016volumetric}, and the safe set represented as a convex corridor of axis-aligned boxes (Fig. \ref{fig:planning}, left). Additionally, we build on top of Gymnasium and Stable Baselines \cite{towers2024gymnasium, stable-baselines3} for easy construction of vectorized environments on separate threads, and for future use in training reinforcement-learning-based policies.

\section{Planning}
\label{sec:planning}

\begin{figure}
  \begin{center}
    \includegraphics[width=\linewidth]{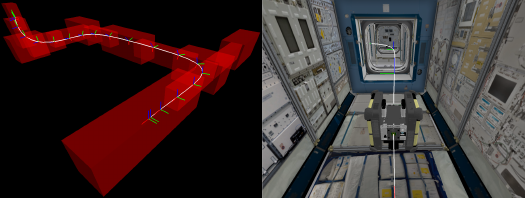}
  \end{center}
  \caption{\textbf{Trajectory planning and tracking.} Left: The \textit{global planner} provides minimum-jerk trajectories (white) through the convex-corridor safe set (red). Right: a snapshot of Astrobee tracking a reference plan with the provided PD tracking controller.} 
  \label{fig:planning}
\end{figure}

For smooth trajectory planning between modules of the ISS, we include a sequential convex programming (SCP)-based global planning method (Fig. \ref{fig:planning}) to determine a time-optimal minimum-jerk B\'ezier spline trajectory through the ISS, building on previous Astrobee planners \cite{gusto, nngusto, watterson_traj_gen} and recent methods \cite{fastpathplanning}. With this, we plan globally-optimal trajectories that enforce all key constraints: guaranteeing that the plan remains in the safe set, continuity and smoothness of the curve and its derivatives, and adhering to the dynamic limits of Astrobee's actuators and operating flight profile. For orientation planning, we use fifth-order quaternion polynomials with boundary conditions on angular velocity and acceleration \cite{quaternion_polynomials}. 

The time to compute the global plan is on the order of 5 to 10 seconds -- reasonable for an initial plan, yet infeasible for online computation and dynamic replanning. To address this, we include a \textit{local planner}, to rapidly (1 ms) compute a single B\'ezier curve and quaternion polynomial which enforces the two-point boundary problem between the robot's current dynamics and a desired terminal state. 

Additionally, \textit{pyastrobee} provides trajectory planners for simple face-forward maneuvers, reorientations, and multi-Astrobee coordinated maneuvers.

\section{Cargo Modeling}

\begin{figure}
  \begin{center}
    \includegraphics[width=0.9\linewidth]{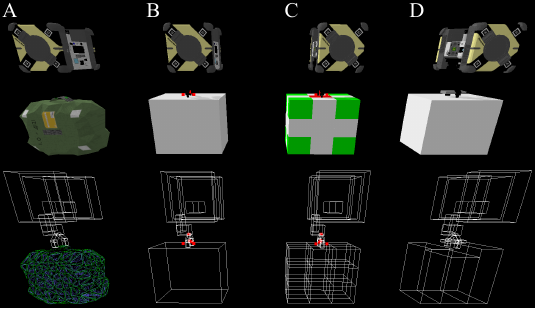}
  \end{center}
  \caption{\textbf{Modeling deformable cargo.} We provide models of the deformable cargo of varying fidelity, including a finite-element deformable bag (A), soft-constraint-handle bag (B), composite-body bag (C), and a URDF model (D), as shown with their wireframe views. Additional models are provided for different handle locations, and multiple handles (for multi-robot manipulation).}
  \label{fig:modeling}
\end{figure}

Accurately modeling the dynamics of a deformable object in simulation remains an open challenge: no model will be able to perfectly match the true behavior, and there is a trade-off between computational cost and accuracy. Given this, we provide four models of the cargo bag (Fig. \ref{fig:modeling}) to use in the cargo transport task. The highest-fidelity bag is represented as a deformable volumetric mesh (via Bullet's FEM deformable capabilities), then the ``constraint" and ``composite" bags simplify the bag as a rigid body (or set of rigid bodies) with a flexible attachment to the Astrobee's gripper using Bullet's soft constraints. Lastly, we provide a URDF, which is the simplest and fastest to analyze, but cannot reflect high deformations of the handle. Further analysis is required to compare these models with the true deformation and dynamics of the CTB.

\section{Control}

\begin{figure}
  \begin{center}
    \includegraphics[width=\linewidth]{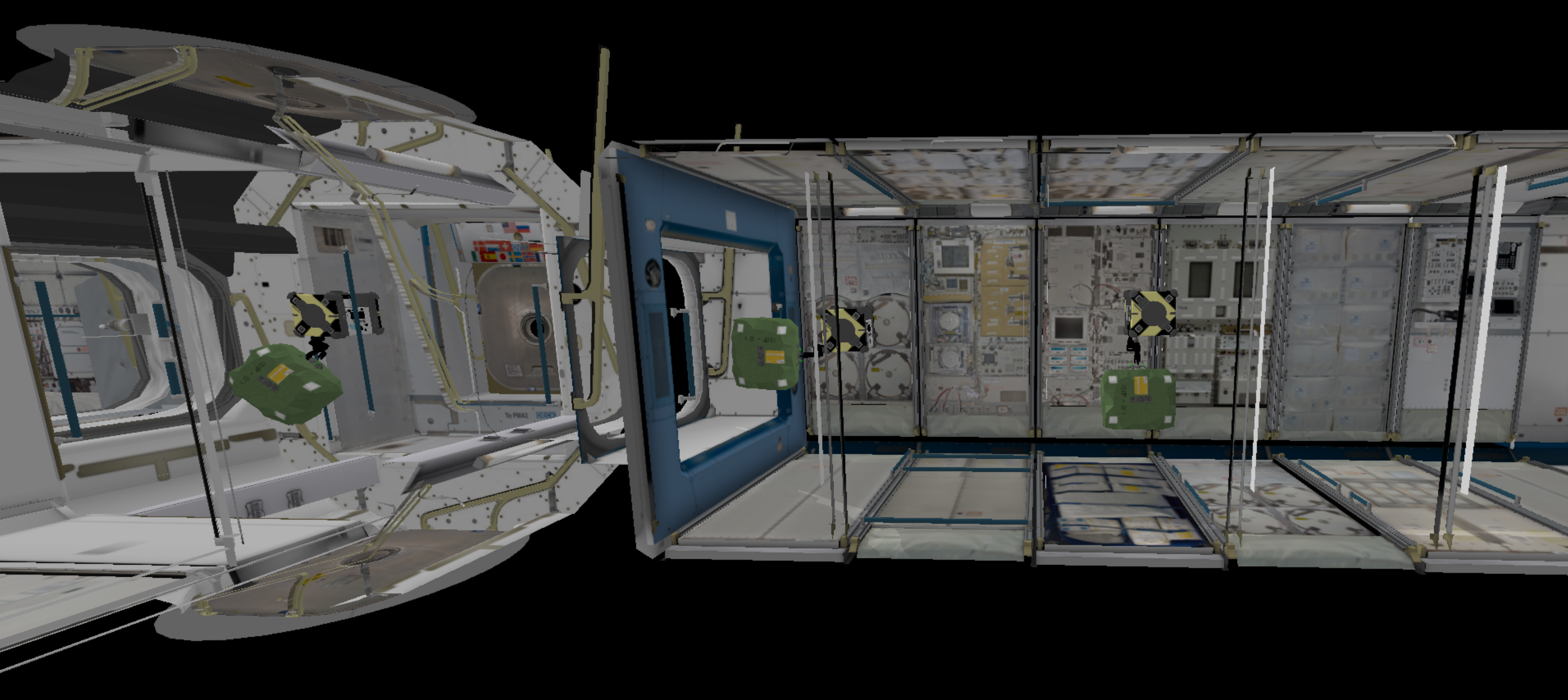}
  \end{center}
  \caption{\textbf{Controlling the cargo transport task.} Our preliminary sampling-based MPC stabilizes the system while transporting the bag through a tight corridor between the Node 2 and JEM modules, without collision.}
  \label{fig:control}
\end{figure}

For controlling Astrobee, we provide two main methods: a simple PD trajectory-tracking controller, and a sampling-based MPC for the cargo manipulation task. Astrobee, when the arm is stowed and no payloads are attached, benefits from its point-robot double-integrator dynamics, making mobility tasks simple with just a PD controller (visualized in Fig. \ref{fig:planning}). However, this approach does not work for the coupled and underactuated dynamics of the Astrobee/CTB system. After grasping the cargo, if the long-horizon effects of the control are not considered, the cargo tends to drift or swing into the walls of the ISS --- particularly, when maneuvering through the narrow corridors between modules and around corners.

As a preliminary approach to solving this problem, and to account for the longer-horizon effects of a control action, we present a sampling-based MPC along the lines of recent work in sampling-based predictive control \cite{mppi, howell2022predictive, rubinstein1999cross, kurtz2025generative}. Since we lack a closed-form model of the deformable CTB's dynamics, we use the simulator as the model\footnote{Note: making the simulator work as the model required a modification to the Bullet source to improve the {\small\texttt{saveState}/\texttt{restoreState}} mechanic for deformable objects.}. By launching multiple simulations on separate threads, we can roll out the effects of a perturbed control sequence (determined by the \textit{local planner}), and take the first \(n\) actions from the best control sequence in a receding-horizon fashion. Each parallelized simulation thread operates on a reduced-order model of the cargo bag, while the primary simulation operates on the full-order FEM deformable bag. We design the cost function to penalize collisions, relative velocities between the CTB and Astrobee, and tracking error to the nominal reference trajectory.

\section{Conclusion} 
\label{sec:conclusion}

We aim for \textit{pyastrobee} to be a valuable resource for those interested in space robotics, manipulation, planning, and control. There are many areas for future work though: particularly, computational efficiency, safety, and multi-robot control. The sampling-based MPC presented is currently too slow to run in real-time, and a closed-form reduced-order model that can be directly integrated without requiring simulation in the loop will likely perform well, even if the model is approximate. Adding a high-frequency CBF safety filter layer (such as via \cite{morton2024cbfpy, morton2025oscbf}) to the control stack would also help maintain collision avoidance guarantees, especially if the MPC is running at a low frequency. Additionally, if two Astrobees can be used for cargo transport (one holding either side of the CTB), this will help mitigate some of the inherent instability of the system. 

\newpage

\section{Acknowledgments}

Daniel Morton was supported by a NASA Space Technology Graduate Research Opportunity.

\bibliographystyle{unsrtnat}
\bibliography{references}

\end{document}